%% file: paper.ijcb2025.fantasyIDiap.tex
\documentclass[10pt,twocolumn,letterpaper]{article}

\usepackage{cvpr} 
\usepackage[dvipsnames]{xcolor}
\usepackage{graphicx}
\usepackage{amsmath}
\usepackage{amssymb}
\usepackage{pifont}
\usepackage{booktabs}
\usepackage[utf8]{inputenc}
\usepackage[T1]{fontenc}
\usepackage{times}
\usepackage[hyphens]{url}
\usepackage{siunitx}
\usepackage{todonotes}
\usepackage{multirow}

\usepackage[pagebackref=true,breaklinks=true,letterpaper=true,colorlinks,bookmarks=false]{hyperref}

\usepackage[capitalize]{cleveref}
\crefname{section}{Sec.}{Secs.}
\Crefname{section}{Section}{Sections}
\Crefname{table}{Table}{Tables}
\crefname{table}{Tab.}{Tabs.}

\newcommand{\cmark}{\textcolor{PineGreen}{\ding{51}}}%
\newcommand{\xmark}{\textcolor{BrickRed}{\ding{55}}}%

\usepackage{caption}
\graphicspath{{figures/}}
\DeclareGraphicsExtensions{.eps,.png,.pdf,.jpg,.JPG}
 
\begin{document}

\title{FantasyID: A dataset for detecting digital manipulations of ID-documents}

\author{
Pavel Korshunov,
Amir Mohammadi,
Vidit,
Christophe Ecabert, and
S\'{e}bastien Marcel \\
Idiap Research Institute, Martigny, Switzerland\\
{\tt\small \{pavel.korshunov, amir.mohammadi, vidit.vidit, christophe.ecabert, sebastien.marcel\}@idiap.ch}}
\maketitle
\thispagestyle{empty}

\input{sections/abstract}


\input{sections/introduction}

\input{sections/related_work}

\input{sections/dataset}

\input{sections/results}

\input{sections/conclusion}

\section*{Acknowledgement}
This work was funded by InnoSuisse 106.729 IP-ICT.

{\small
\bibliographystyle{ieee}
\bibliography{references}
}

\end{document}

%% file: sections/abstract.tex
\begin{abstract}

Advancements in image generation led to the availability of easy-to-use tools for malicious actors to create forged images. 
These tools pose a serious threat to the widespread Know Your Customer (KYC) applications, requiring robust systems for detection of the forged Identity Documents (IDs). To facilitate the development of the detection algorithms, in this paper, we propose a novel publicly available (including commercial use) dataset, FantasyID, which mimics real-world IDs but without tampering with legal documents 
and, compared to previous public datasets, it does not contain generated faces or specimen watermarks. 
FantasyID contains ID cards with diverse design styles, languages, and faces of real people. To simulate a realistic KYC scenario, the cards from FantasyID were printed and captured with three different devices, constituting the bonafide class. We have emulated digital forgery/injection attacks that could be performed by a malicious actor to tamper the IDs using the existing generative tools.
The current state-of-the-art forgery detection algorithms, such as TruFor, MMFusion, UniFD, and FatFormer, are challenged by FantasyID dataset. It especially evident, in the evaluation conditions close to practical, with the operational threshold set on validation set so that false positive rate is at 10\%,  leading to false negative rates close to 50\% across the board on the test set. 
The evaluation experiments demonstrate that FantasyID dataset is complex enough to be used as an evaluation benchmark for detection algorithms. 
\end{abstract}

%% file: sections/introduction.tex
\input{figures/teaser_fig}

\input{figures/other_cards}
\section{Introduction}
\label{sec:intro}

Different financial services, like banks and insurance companies, use face and document verification systems to authenticate their users. This digital \emph{Know Your Customer} (KYC) lets users take pictures of their ID (passports, residence permits, and driving licenses) using phone camera. The captured image is then typically compared to a selfie of the user to validate the authenticity of the document and the user's information. This process results in quick onboarding and increases the efficiency of the overall system. 

However, such KYC process presents a security risk, as one can use fake or forged documents to create a fraudulent account. The malicious user can potentially bypass the camera capture API and directly \emph{inject} a forged image for authentication. Alternatively, an attacker can print the digitally forged document, they have access to ID documents printing equipment, and capture it using a `normal' onboarding process thus fooling the system. This vulnerability of KYC systems is exacerbated by the availability of rapidly improving image generation and editing methods~\cite{ji_improving_2023,chen2025textdiffuser,chen_simswap_2024,rosberg_facedancer_2023} allowing realistic-looking forged documents to be generated within minutes. To mitigate these risks, KYC process requires appropriate detection algorithms to flag forged documents.

There is a lack of publicly available datasets suitable for the detection of forged ID documents. The main reason is the restriction by official authorities on tampering of legal documents and publicizing sensitive personal information. Previous public ID datasets~\cite{arlazarov_midv-500_2019,bulatov_midv-2019_2020,bulatov_midv-2020_2022,boned_synthetic_2024,al-ghadi_guilloche_2023,park_kid34k_2023} are either not meant for the manipulation-detection task, lack diversity, or contain bonafide samples that are tampered versions of official specimen ID documents (the watermark is manually or automatically removed). 

To mitigate these issues, we introduce the first publicly available for commercial and non commercial use dataset, FantasyID\footnote{\url{https://www.idiap.ch/paper/fantasyid}}, that contains both genuine and forged ID cards. Our ID cards (\cref{fig:teaser}) are designed to resemble official ID documents (such as passports or ID cards) while avoiding legal problems associated with official ID document tampering, hence the name \emph{Fantasy ID}. We created $13$ templates representing genuine/bonafide IDs in the unique style of the following languages: Arabic, Chinese, Hindi, French, Persian, Portuguese, Russian, Turkish, Ukrainian, and English. All designs were crafted using Creative Commons $4.0$-licensed source materials. The ID cards do not strictly follow the ICAO~\cite{ICAO9303_MRTD} standard for travel documents, because we used face images with public access licenses, so they do not follow strict passport-photo standards and because these cards are not meant to serve as official documents. However, our ID cards contain the main elements of an ID document, such a facial image, different text, including official and personal information, design elements resembling some of the real-world documents and Guilloche patterns often present in the ID documents.

FantasyID dataset has the following unique and novel characteristics:
\begin{enumerate}
    \item FantasyID is the first public dataset where the \emph{bonafide} cards are not modified versions of some official ID cards (e.g., with a digitally removed word `specimen'). We provide pristine \emph{bonafide} cards, which is important, since tampered images will bias digital manipulations detection algorithms.
    \item FantasyID contains IDs for several non-English languages and is created using design patterns that mimic the styles used in the corresponding cultures. Hence, the FantasyID dataset facilitates research in multi-lingual text manipulation detection.
    \item We use the faces of real people to avoid biasing our \emph{bonafide} to generated fake faces. 
    \item In addition to \emph{bonafide} cards, we provide digital manipulations created using face swapping approaches, such as InSwapper and Facedancer~\cite{rosberg2023facedancer}, and text inpainting techniques, such as Textdiffuser2~\cite{chen2025textdiffuser} and DiffSTE~\cite{ji_improving_2023} (see more details in \cref{sec:creation}).
\end{enumerate}

In the following sections, we discuss the existing datasets of ID documents, the generative methods used to create forged documents, and the existing forgery detection approaches; describe the process of building the FantasyID dataset; and present evaluation results of baseline forgery detection algorithms to demonstrate how challenging and useful FantasyID is for forgery detection research.

%% file: figures/teaser_fig.tex
\begin{figure}[htb]
\centering
	\begin{subfigure}{.49\textwidth}
		\includegraphics[width=\linewidth]{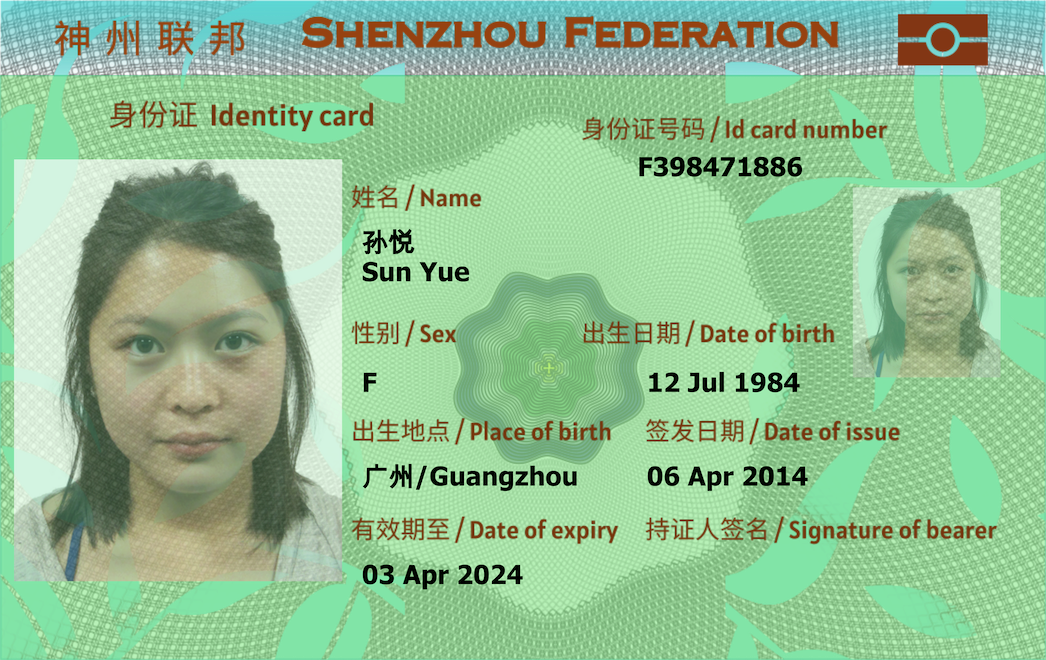}
		\caption{Chinese language}
	\end{subfigure}
	~	
	\begin{subfigure}{.49\textwidth}
		\includegraphics[width=\linewidth]{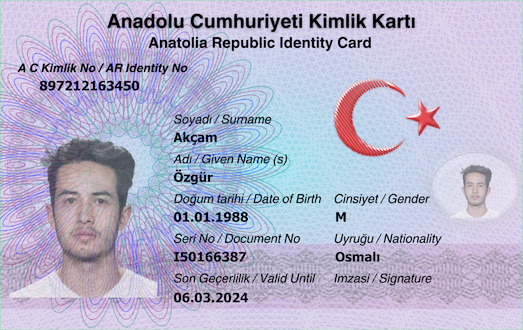}
		\caption{Turkish language}
	\end{subfigure}
\caption{\textbf{FantasyID:} Examples of original digital versions of FantasyID cards. The faces are of real people but the other biometric information is not real. The cards contain Guilloche patterns and design elements inspired by the ones used in  official ID documents. We design $13$ different card templates which are then physically printed with biometric details and recaptured to create $1086$ \emph{bonafide} images.  }
\label{fig:teaser}
\end{figure}

%% file: figures/other_cards.tex
\begin{figure*}[htb]
\centering
\subfloat[Persian language]{\includegraphics[width=0.25\textwidth]{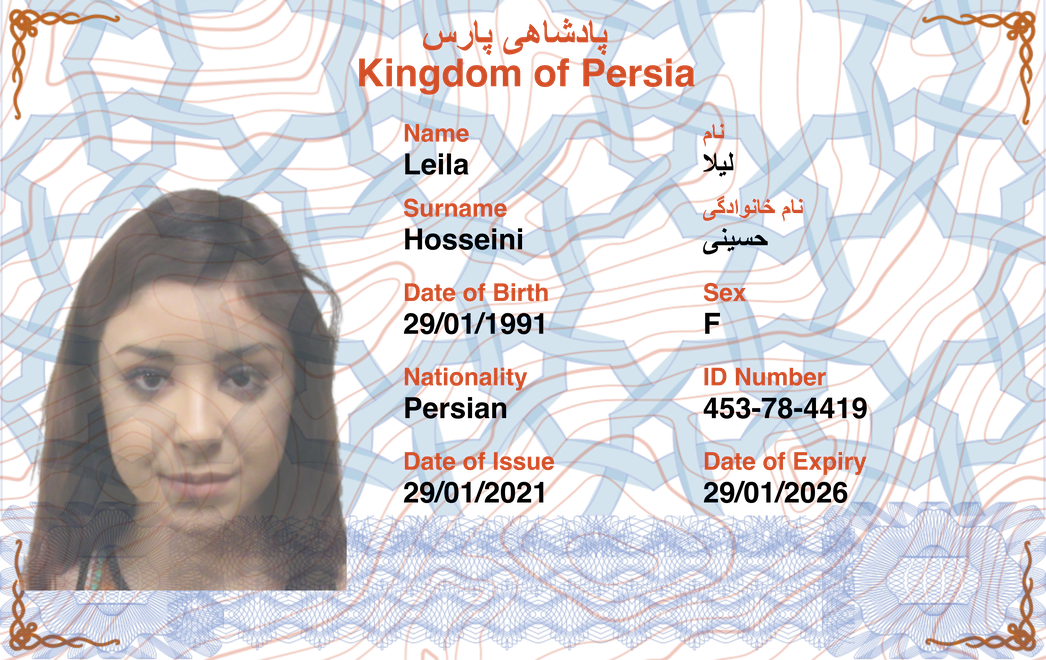}}
\subfloat[French language]{\includegraphics[width=0.25\textwidth]{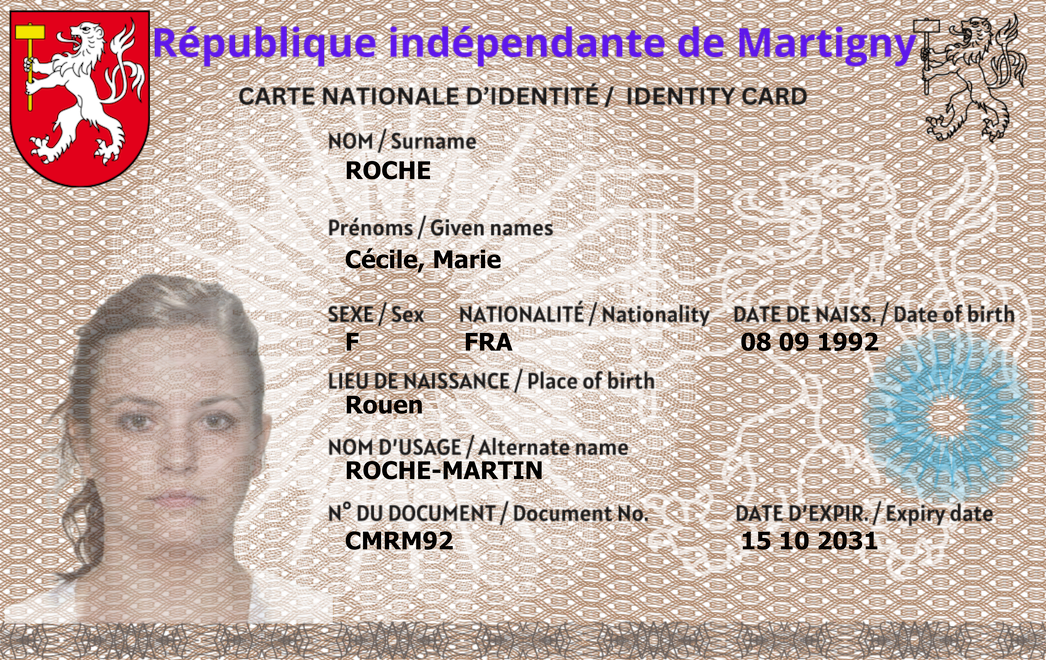}}  
\subfloat[Arabic language]{\includegraphics[width=0.25\textwidth]{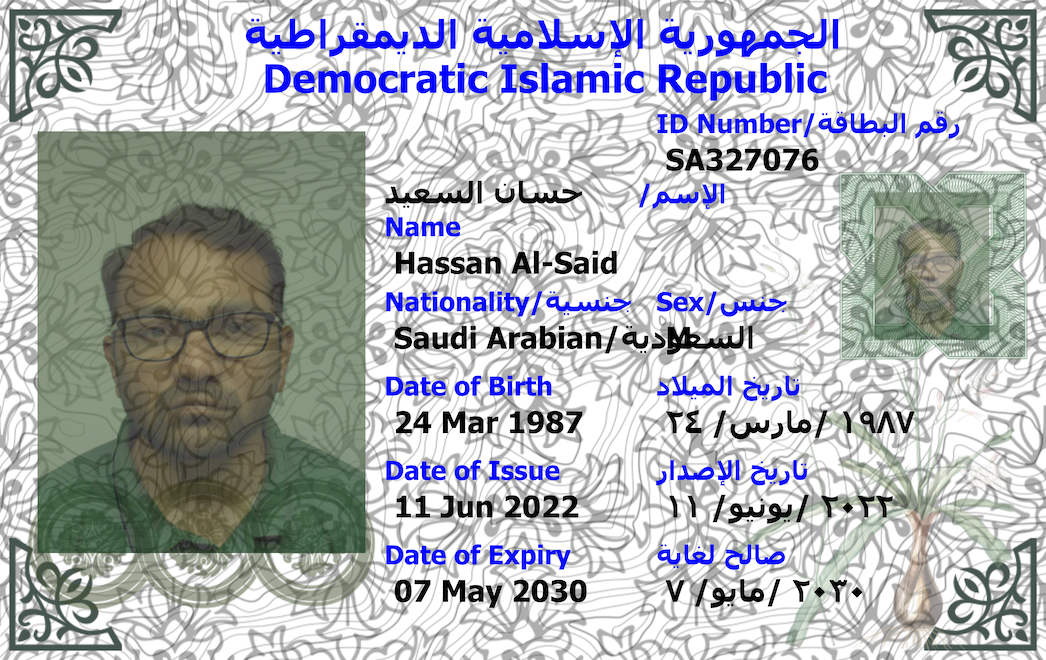}}  
\subfloat[Russian language]{\includegraphics[width=0.25\textwidth]{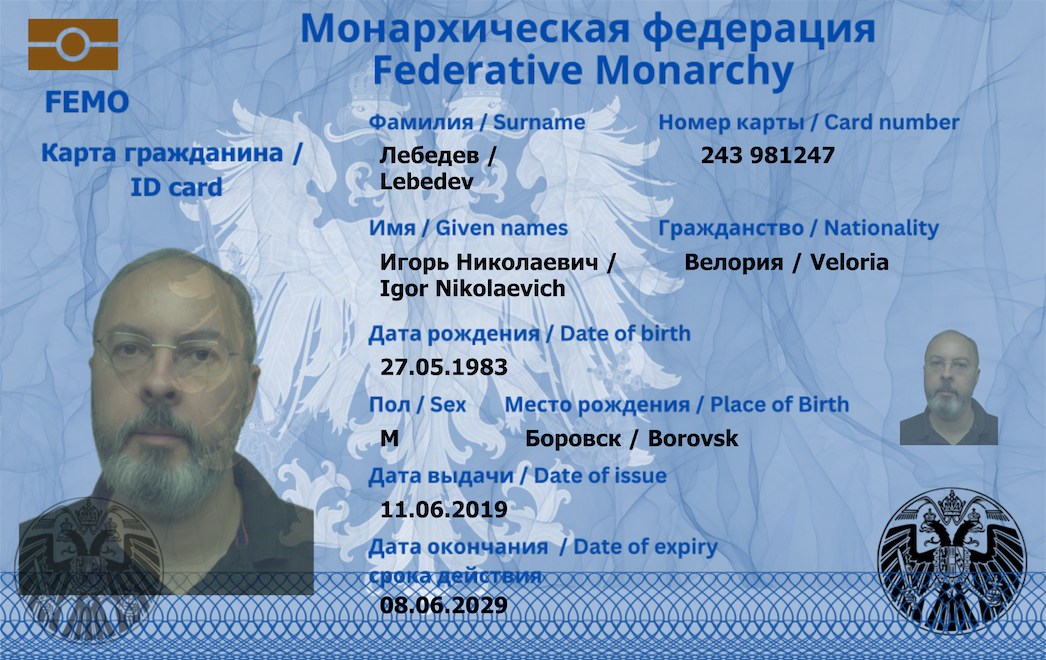}}  
\caption{Other examples of original digital versions of FantasyID cards.}
\label{fig:other_cards}
\end{figure*}

%% file: sections/related_work.tex
\begin{figure*}[htb]
\centering
\subfloat[Indian, iPhone $15$ Pro]{\includegraphics[width=0.49\textwidth]{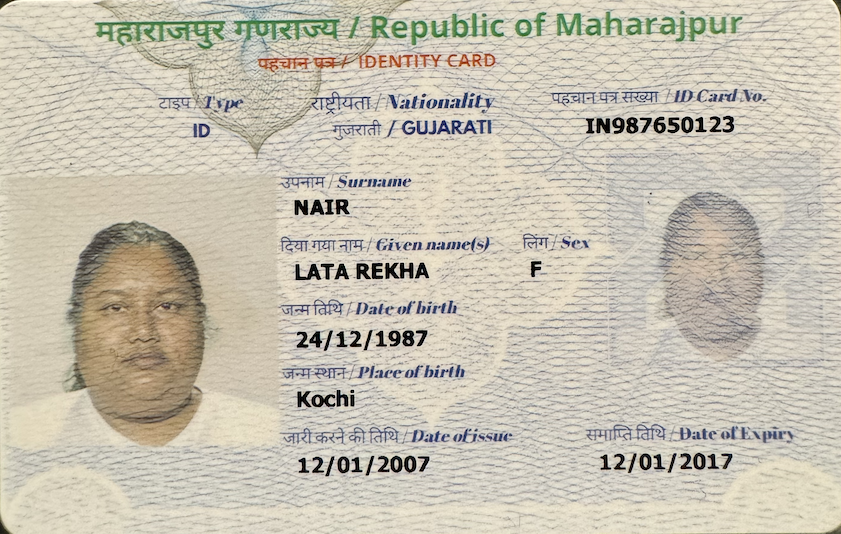}}
~
\subfloat[Ukrainian, Huawei Mate $30$]{\includegraphics[width=0.49\textwidth]{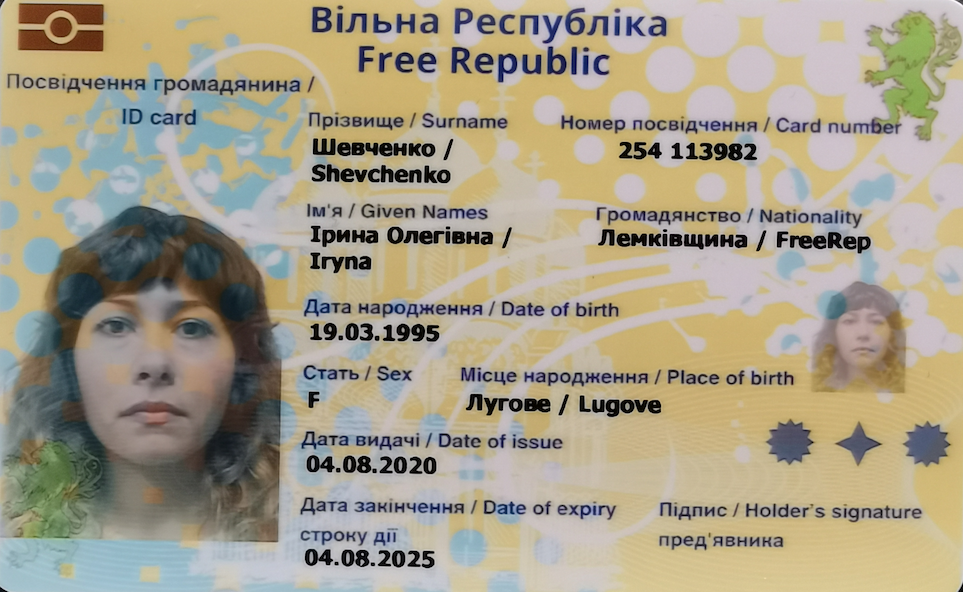}}  
\caption{\textbf{Printed Bonafide:} Examples of bonafide ID cards printed and captured with different devices.}
\label{fig:bfcaptured}
\end{figure*}

\section{Related Work}
We summarize different ID datasets (\cref{tab:id_datasets}) highlighting their intended purposes and evaluating their suitability for manipulation detection tasks.

\paragraph{MIDV-500~\cite{arlazarov_midv-500_2019}.} This dataset is based on $50$ specimen copies of the real IDs sourced from Wikimedia Commons\footnote{\url{https://commons.wikimedia.org/wiki/Main\_Page}}, with the term ``specimen'' digitally removed. The images were printed and then the video was captured using different devices and backgrounds. The dataset was created to recognize and analyze the IDs using mobile devices.
The tampering of the originals makes them biased and they cannot be considered as \emph{bonafide}.  

\paragraph{MIDV-2019~\cite{bulatov_midv-2019_2020}.} It extends MIDV-500 by capturing the physical cards under large perspective distortion and low-light conditions. The primary use case of this dataset is similar to MIDV-500 for ID recognition and analysis. It also has the same tampering bias in the \emph{bonafide} images.

\paragraph{MIDV-2020~\cite{bulatov_midv-2020_2022}.} This dataset digitally alters texts and faces of the $10$ IDs in MIDV-500 to create $1000$ images. The primary goal is to study the text and face region detection using this dataset. The authors propose to use these $1000$ images as fake documents to evaluate forgery detection algorithms. However, the original MIDV-500  manipulated regions like the removed word "specimen" are not labeled. This creates a bias for the detection algorithms.

\paragraph{FMIDV~\cite{al-ghadi_guilloche_2023}.}  It was proposed to detect guilloche pattern-based forgeries as they are a common security feature in official ID documents. This dataset is created by applying a copy-move attack on MIDV-2020 images but only in the non-text/face region. These kinds of forgery are interesting but their scope is limited as they do not cover biometrics like faces and texts. Additionally, they treat MIDV-2020 images as \emph{bonafide}, which already has digital manipulations.

\input{figures/real_forged_fig}

\input{tables/listofdatasets}

\paragraph{KID34k~\cite{park_kid34k_2023}.} This dataset was specifically created for online identity card fraud detection. It contains images of digitally created $82$ ID cards based on Korean driver's licenses and registration cards. Here, the biometric information used is fake for both text and faces. The main purpose of the dataset is to study \emph{screen} and \emph{paper} attack detection. Around $34662$ images are created by recapturing images displayed on screens and printed using $2$ kinds of paper printers. The dataset is limited in the diversity of style and language (\emph{only} Korean ) used, which is needed for the robust study of detection algorithms. Moreover, fake faces used to create \emph{bonafide} leads to bias in the dataset.

\paragraph{BID~\cite{soares_bid_2020}.} BID dataset was proposed for ID analysis tasks, such as text and image region segmentation, optical character recognition (OCR), and ID recognition. To create the dataset, biometrics features such as personal details and faces were erased and fake details were added digitally to the original Brazilian ID cards. These changes lead to a corrupted \emph{bonafide}, rendering the dataset not suitable for manipulation detection tasks.

\paragraph{SIDTD~\cite{boned_synthetic_2024}.} Another extension of MIDV-2020 dataset. The images from MIDV-2020 are considered \emph{bonafide} and are used to generate \emph{copy-replace} and \emph{inpainting} based forgeries. These images are physically printed to create a \emph{presentation} attack scenario. As mentioned above, employing MIDV-2020 images as \emph{bonafide} is already biasing the system towards manipulated \emph{bonafide}. 

\paragraph{PAD~\cite{tapia_first_2024}} Recently proposed competition on PAD evaluated several methods with their private database of 300K ID cards. This database consists of \emph{print}, \emph{screen} and \emph{composite} attacks. Since, we specifically study \emph{injection} attack using digital image manipulation, comparison is out of scope with this work.

%% file: figures/real_forged_fig.tex
\begin{figure*}[htb]
\centering
\subfloat[Bonafide English, captured with iPhone $15$ Pro]{\includegraphics[width=0.49\textwidth]{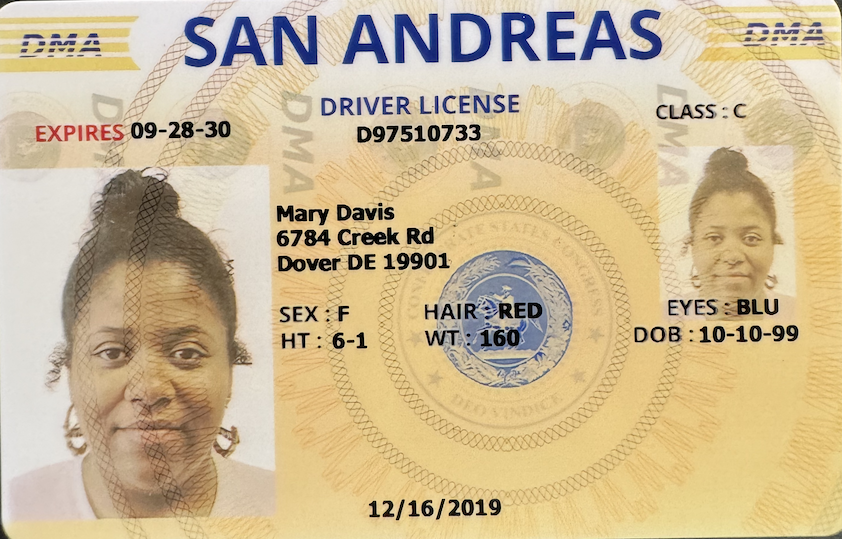}}
~
\subfloat[Digitally manipulated]{\includegraphics[width=0.49\textwidth]{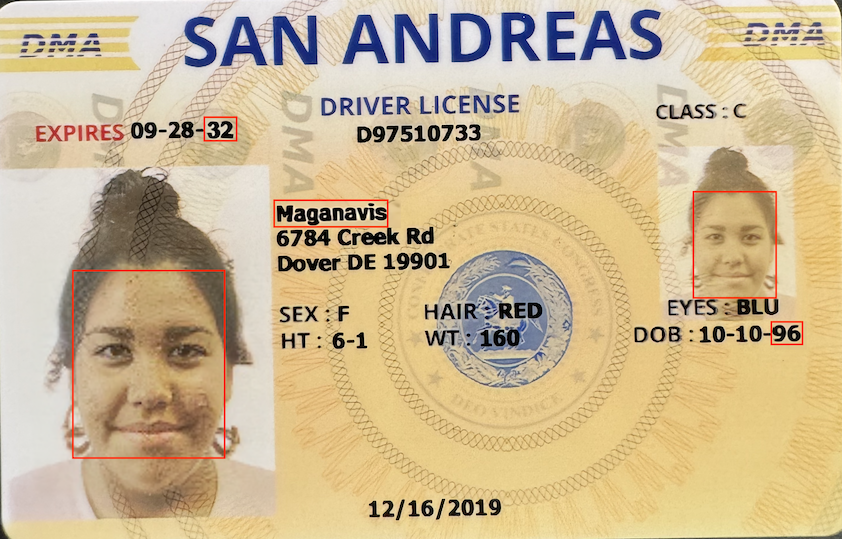}}  
\caption{\textbf{Digital Forgery:} English language ID-card printed and captured with iPhone $15$ Pro, and then manipulated by swapping face and altering text. Note that the box in \textcolor{red}{red} shows digitally manipulated areas. }
\label{fig:digitalfake}
\end{figure*}

%% file: tables/listofdatasets.tex
\begin{table*}[t!]
\centering
\begin{tabular}{@{}lccc|ccc|cc@{}}
\toprule
& & & & \multicolumn{3}{c|}{Bonafide} & \multicolumn{2}{c}{Attack} \\
Dataset & \# IDs & Use Case & Source DB & Real Face & Pristine & Pr. Capt.
& Digit. Manip. & Pr. Capt.  \\
\midrule
MIDV-500~\cite{arlazarov_midv-500_2019} & $50$ & ID Rec. & Web & \cmark & \xmark & \cmark & \xmark & \xmark \\
MIDV-2019~\cite{bulatov_midv-2019_2020} & $50$ & ID Rec. & MIDV-500 & \cmark & \xmark & \cmark & \xmark & \xmark \\
MIDV-2020~\cite{bulatov_midv-2020_2022} & $1000$ & ID Rec.  & MIDV-500 & \xmark & \xmark & \cmark & \cmark & \xmark \\
FMIDV~\cite{al-ghadi_guilloche_2023} & $1000$ & Guilloche Det. & MIDV-2020 & \xmark  & \xmark & \cmark & \cmark & \xmark \\
KID34k~\cite{park_kid34k_2023} & $82$ & PAD & Generated & \xmark & \xmark & \cmark & \xmark & \xmark \\
BID Dataset~\cite{soares_bid_2020} & NA & ID Rec. & Real IDs & \xmark  & \xmark & \xmark & \xmark & \xmark \\
SIDTD~\cite{boned_synthetic_2024} & $10$ & PAD  & MIDV-2020 & \xmark  & \xmark & \cmark & \cmark & \cmark \\
\midrule
FantasyID (Ours) & $362$ & Man. Det. & Generated & \cmark & \cmark & \cmark & \cmark & \cmark \\
\bottomrule
\end{tabular}
\caption{\textbf{ID Datasets.}. The table shows key characteristics of publicly available datasets for ID document analysis. Columns: {\bf Real Face} indicates whether a real face was used on ID, {\bf Pristine} means genuine bonafide cards were not manipulated in any way, {\bf Pr. Capt.} indicates if the cards were printed and captured with a device's camera, and {\bf Digit. Manip.} shows if  cards were digitally manipulated to simulate a forgery attack. Only FMIDV was proposed for manipulation detection but restricted to manipulation in the guilloche patterns.  Except for KID34k and BID, others are based upon MIDV-500 images and carry the bias of the manipulation in \emph{bonafide} images. KID34k is limited in the diversity of style and language used on the cards. Furthermore, it uses generated faces for \emph{bonafide} which will bias the detection algorithms. Our, FantasyID,  provides pristine \emph{bonafide} images with real faces along with diversity in card design and languages present. ID Rec. stands for ID Recognition and Man. Det. for Manipulation Detection. PAD is Presentation Attack Detection.}
\label{tab:id_datasets}
\end{table*} 


%% file: sections/dataset.tex
\section{FantasyID Dataset}
\label{sec:dataset}

In this section, we provide an overview of the process of designing and building the \emph{FantasyID} dataset, a diverse set of identity cards aimed at advancing research in forgery detection algorithms within biometrics. Our dataset consists of two main categories of ID cards: \emph{bonafide}, which are the pristine cards, and \emph{attacks}, which are different \emph{digital} manipulation of pristine cards.

\subsection{Creation of Bonafide Fantasy ID Cards}
The first category of the dataset focuses on genuine \emph{bonafide} fantasy ID cards. This process involved three key steps: ID card design and generation, printing, and image capture, respectively. We assume the scenario when a genuine user will capture their genuine physical card with a phone or scanner when enrolling into KYC service.

\input{tables/fantasydatasets}

\begin{enumerate}
    \item \textbf{ID Card Generation}: We generated $262$ genuine ID cards for training set and additional $100$ ID cards for test set, using random personal data (e.g., date of birth, names, city of origin) specific to each language (see \cref{fig:teaser} and \cref{fig:other_cards} for some examples). These cards feature real faces from datasets such as the American Multiracial Face Database (AMFD)~\cite{chen2021broadening}, Face London Research Dataset~\cite{DeBruine2017},  High-Quality Wide Multi-Channel Attack (HQ-WMCA)~\cite{mostaani2020high} and a set of $100$ face images with open license (public domain or CC-BY-4.0) manually downloaded by us from Flickr.

    \item \textbf{Printing}: The generated cards were printed using Evolis Primacy 2 card printer\footnote{\href{https://www.evolis.com/solutions/card-printers/primacy2-card-printer/}{Evolis Primacy 2 Printer}} to emulate real-world ID cards. This step resulted in $362$ high-quality printed ID cards \emph{($600$ DPI)}, prepared for further processing.    
    
    \item \textbf{Image Capture}: Digital images of each printed ID card were captured using three devices (Apple iPhone $15$ Pro, Huawei Mate $30$, and Kyocera TASKalfa $2554$ci office scanner). See~\cref{fig:bfcaptured} for the examples of printed and captured \emph{bonafide} cards. A set of the $362$ printed cards was captured three times (once with each device) resulting in a total of $1086$ \emph{($362 \times 3$)} high-quality images of \emph{bonafide} fantasy ID cards. We use $786$ bonafide cards for train-val set and remaining $100$ for testing.
\end{enumerate}

\subsection{Creation of Manipulated Fantasy ID Cards}
\label{sec:creation}
The second category of the dataset deals with fake fantasy ID cards. These manipulated cards form the digitally altered attacks of FantasyID and represent the use case when an attacker is trying to subvert a KYC system by submitting a digital fake version of an ID card with face, name, or/and date of expiry altered (see an example in \cref{fig:digitalfake}). The presentation of this fake card is done digitally by bypassing a sensor of the KYC system. This category is further divided into two sub-categories based on manipulation methods: 

\begin{enumerate}
	\item \textbf{Train-Val}: $786$ bonafide cards which are captured using three different devices  
     were digitally modified by swapping faces and inpainting text regions. We create two set of manipulations by using different swapping and inpainting methods, namely: (a) InSwapper (InsightFace\footnote{\url{https://insightface.ai/}}) for face and DiffSTE~\cite{ji_improving_2023} for text
     (b) Facedancer~\cite{rosberg2023facedancer} for face and Textdiffuser2~\cite{chen2025textdiffuser} for text. \cref{fig:digitalfake} demonstrates the results of these digital manipulations. Therefore, we have $786$ bonafide and $1572$ manipulated images in this set. We keep all $459$ cards containing faces from HQ-WMCA dataset as val set and the rest of $1899$ images as training set. The training set can be used to fine-tune a detection model. 

    \item \textbf{Test}: In order to create a challenging test set, we create distinct manipulations on the bonafide of train-val set, i.e., \textbf{Attack-1}. It consists of  $786$ images with \emph{text-only} modification created by finetuning Textdiffuser2~\cite{chen2025textdiffuser}. The generated text regions are post-processed with the Segment Anything Model~\cite{kirillov2023segment} to extract text regions with their background, and alpha blended with original background, where the text region is erased using LaMA-inpaint~\cite{suvorov2021resolution}. This approach creates text modifications that are hard to notice visually.

    Further to test the out-of-domain generalization, we create a bonafide set consisting of $300$ images which are distinct from train-val set in terms of template design, faces, and textual details.
    We modify these $300$ cards using different kind of manipulations: (b) \textbf{Attack-2} consists of \emph{face-only} manipulation using Facedancer~\cite{rosberg2023facedancer} on $150$ ID cards and (c) \textbf{Attack-3} is created by changing \emph{text-only} regions using the same approach as in Attack-1 but applying it to the $149$ ID cards from the test set. In total, we have $300$ bonafide and $1085$ manipulated images in the test set. 

\end{enumerate}

Each manipulation category aims to reflect real-world attack scenarios for testing the robustness of detection algorithms. \cref{tab:fantasydatasets} summarizes the key steps, categories, devices, and purposes involved in the creation of genuine and fake FantasyID cards.

%% file: tables/fantasydatasets.tex
\begin{table*}[t]
\renewcommand{\arraystretch}{1.2}
\centering
\begin{tabular}{ p{0.12\textwidth}  p{0.17\textwidth}  c  p{0.2\textwidth}  p{0.34\textwidth} }
\hline
\textbf{Category} & \textbf{Sub-Category} & & \textbf{Devices/Sources} & \textbf{Description and Purpose} \\
\hline

\multirow{3}{*}{Bonafide Cards} 
  & Generation & $362$ & AMFD, Face London, WMCA, and Flickr datasets & Thirteen unique design styles in ten languages. Random but realistic personal info. \\
\cline{2-5}

& Print & $362$ & Evolis Primacy 2 printer & Printed on physical plastic cards. \\
\cline{2-5}
& Capture & $1086$ &  iPhone $15$ Pro, Huawei Mate $30$, office scanner & Plastic cards were captured using three devices. \\
\hline

\multirow{4}{*}{Forged Cards} 
& Digital manipulation & $786$ & InSwapper & Each face in captured IDs from \emph{train-val} set is swapped with another face. \\
\cline{2-5}
& Digital manipulation & $786$ & Facedancer~\cite{rosberg2023facedancer} & Each face in captured IDs from \emph{train-val} set is swapped with another face. \\
\cline{2-5}
& Digital manipulation (Attack-2) & $150$ & Facedancer~\cite{rosberg2023facedancer} & Faces in captured IDs from subset of \emph{test} set are swapped. \\
\cline{2-5}
& Digital manipulation & $786$ & DiffSTE~\cite{ji_improving_2023} & Parts of personal info in captured IDs from \emph{train-val} set were replaced by another text. \\
\cline{2-5}
& Digital manipulation & $786$ &Textdiffuser2~\cite{chen2025textdiffuser} & Parts of personal info in captured IDs from \emph{train-val} set were replaced by another text. \\
\cline{2-5}
& Digital manipulation (Attack-1) & $786$ & Finetuned-Textdiffuser2 & Parts of personal info in captured IDs from \emph{train-val} set were replaced by another text. \\
\cline{2-5}
& Digital manipulation (Attack-3) & $149$ & Finetuned-Textdiffuser2 & Parts of personal info in captured IDs from \emph{test} set were replaced by another text. \\
\cline{2-5}
\hline

\end{tabular}
\caption{The summary of how different parts of FantasyID dataset were created.}
\label{tab:fantasydatasets}
\end{table*}

%% file: sections/results.tex
\section{Evaluation of Baselines on FantasyID}
\label{sec:results}

FantasyID dataset is the first dataset of ID documents where bonafide digital originals and the digital attacks on the face and text data are available in the public domain. Nevertheless, to demonstrate that this dataset also poses a challenge to algorithms designed to detect manipulations and document forgeries, we conducted a set of experiments evaluating baseline algorithms on this dataset. For the baseline algorithms, we have only considered methods that focus on generic synthetic image or local region manipulations detection, as oppose to more known methods of deepfake detection that typically focus on faces~\cite{Anubhav2021,Korshunov2022}.

In this paper, we used the following four state of the art algorithms for binary detection of manipulations in images: TruFor~\cite{guillaro_trufor_2023}, MMFusion~\cite{rudinac_exploring_2024}, UniFD~\cite{ojha_towards_2023}, and FatFormer~\cite{liu_forgery-aware_2024}.
We have used the pretrained models provided by the authors.
The fake images used in the training datasets of TruFor and MMFusion are tampered images where only some parts of the image are modified.
Whereas the fake images seen during training of UniFD and FatFormer are GAN generated images where the full image is digitally generated.
More details about the algorithms are given below.

\begin{itemize}
    \item {\bf TruFor~\cite{guillaro_trufor_2023}}
     uses a multi-branch Transformer encoder architecture to combine features from RGB images and Noiseprint++ images to predict an anomaly localization map, a confidence map, and a final score.
    Noiseprint++~\cite{guillaro_trufor_2023} is a fully convolutional network trained to extract the subtle noise present in pristine images caused by imperfections in camera hardware or in-camera processing steps.
    In our binary detection experiments, we used the final score of TruFor.
    
    \item {\bf MMFusion~\cite{rudinac_exploring_2024}}
     extends TruFor by adding more modalities to the encoder architecture.
    The authors propose to use Steganalysis Rich Model (SRM) filtered and Bayar convolution images in addition to the Noiseprint++ images.
    The integration of SRM and Bayar images is done in two ways: i) early fusion, when images (except for RGB) are passed into a convolution block and are merged into one image before being used as input to the encoder architecture, and ii) late fusion, when the encoder is repeated for each pair of RGB and one other image modality and their final features are combined.
    The weights of the RGB branches are shared in this case.
    In our binary detection experiments, we have used the early fusion variant of MMFusion because of its better performance.
    \item {\bf UniFD~\cite{ojha_towards_2023}}
     is a simple approach that uses features from a pretrained frozen CLIP:ViT-L/14~\cite{radford_learning_2021} model and a linear classifier to detect fake images. The aim of UniFD is to detect fully synthetic images, so it may under-perform on images that are only partially modified.
    \item {\bf FatFormer~\cite{liu_forgery-aware_2024}}
     uses the CLIP:ViT-L/14 model similar to UniFD.
    However, it adds forgery-aware adapters to the ViT model that contains convolution and discrete wavelet transform operations.
    It also introduces language-guided alignment using the text encoder of CLIP to guide the image encoder to focus on forgery-related representations.
\end{itemize}

\subsection{Evaluation Protocol and Metrics}
\label{sec:protocol}

To make sure the performance of the baseline methods does not degrade due to the difference between training and testing, we applied the same preprocessing step to the input images that were proposed by the authors of the corresponding algorithms.
TruFor and MMFusion baselines work on full resolution images.
The input images to UniFD are resized so that the shortest side is $224$ pixels and are then center cropped to obtain a $224\times224$ square image.
The input images to FatFormer are first resized to $256\times256$ pixels and then center cropped to $224\times224$ pixels.
We also evaluated both UniFD and FatFormer on zero-padded images to make the images square but that lead to worse performance.


For evaluation metrics, we report commonly used metrics for binary classification: false positive rate (FPR), where positives are \emph{bonafide} images, false negative rate (FNR), and half total error rate (HTER), which is the average of FPR and FNR. These rates are computed on the \textit{Test} set using a threshold estimated on the \textit{Val} set at `FPR$=10$\%'.
In addition to these metrics, to be comparable with the metrics used by the authors of the baseline methods, we also report area under the curve (AUC) of ROC plots, balanced accuracy (ACC), and F1 score weighted by class on the \textit{Test} set. ACC and F1 are computed using a fixed threshold of $0.5$.

Even though we use the baseline algorithms \textit{as is} without additional tuning on the training set of FantasyID, we evaluate them on the test set only. In this way, FantasyID test set can be used as a benchmark in the future to compare different methods for ID documents manipulation detection, while the train-val set can be used to tune the models. Hence, we compute metrics for the test set overall and for each of the three attacks (see \cref{sec:creation} for details), to have a better understanding about the impact of each type of manipulation on the performance of the baselines.

\subsection{Evaluation Results}

\renewcommand{\arraystretch}{1.2}
\begin{table}[tb]
\footnotesize
\centering
\setlength{\tabcolsep}{5pt}
\begin{tabular}{llrrrrrr}
\toprule
Model & Protocol & ACC & AUC & F1 & FPR & FNR & HTER \\
\midrule
TruFor & all & 65.9 & 93.5 & 80.7 & 4.9 & 62.0 & 33.4 \\
TruFor & Attack-1 & 67.8 & 99.5 & 79.0 & 0.1 & 62.0 & 31.1 \\
TruFor & Attack-2 & 53.5 & 55.8 & 47.6 & 34.7 & 62.0 & 48.3 \\
TruFor & Attack-3 & 67.8 & 99.9 & 55.3 & 0.0 & 62.0 & 31.0 \\
MMFusion & all & 55.1 & 94.4 & 73.7 & 4.0 & 47.7 & 25.8 \\
MMFusion & Attack-1 & 55.2 & 99.8 & 67.0 & 0.1 & 47.7 & 23.9 \\
MMFusion & Attack-2 & 54.5 & 61.5 & 29.8 & 28.0 & 47.7 & 37.8 \\
MMFusion & Attack-3 & 55.2 & 99.1 & 30.0 & 0.0 & 47.7 & 23.8 \\
UniFD & all & 50.0 & 52.0 & 7.7 & 8.3 & 92.7 & 50.5 \\
UniFD & Attack-1 & 50.0 & 54.3 & 12.0 & 7.4 & 92.7 & 50.0 \\
UniFD & Attack-2 & 50.0 & 48.4 & 53.3 & 7.3 & 92.7 & 50.0 \\
UniFD & Attack-3 & 50.0 & 43.5 & 53.5 & 14.1 & 92.7 & 53.4 \\
FatFormer & all & 48.8 & 53.5 & 15.6 & 6.5 & 92.3 & 49.4 \\
FatFormer & Attack-1 & 49.3 & 57.6 & 20.4 & 1.1 & 92.3 & 46.7 \\
FatFormer & Attack-2 & 48.3 & 43.6 & 53.8 & 24.0 & 92.3 & 58.2 \\
FatFormer & Attack-3 & 46.7 & 42.3 & 51.8 & 16.8 & 92.3 & 54.6 \\
\bottomrule
\end{tabular}
\caption{Baselines methods evaluated on the test set of FantasyID, with digital forged ID cards as the negative set. ACC and F1 are computed using 0.5 as threshold while FPR, FNR, and HTER are computed using 10\% FPR threshold from the validation set.}
\label{tab:results-digital}
\end{table}


\begin{figure*}[htb]
\centering
\subfloat[\textit{all} protocol]{\includegraphics[width=0.45\textwidth]{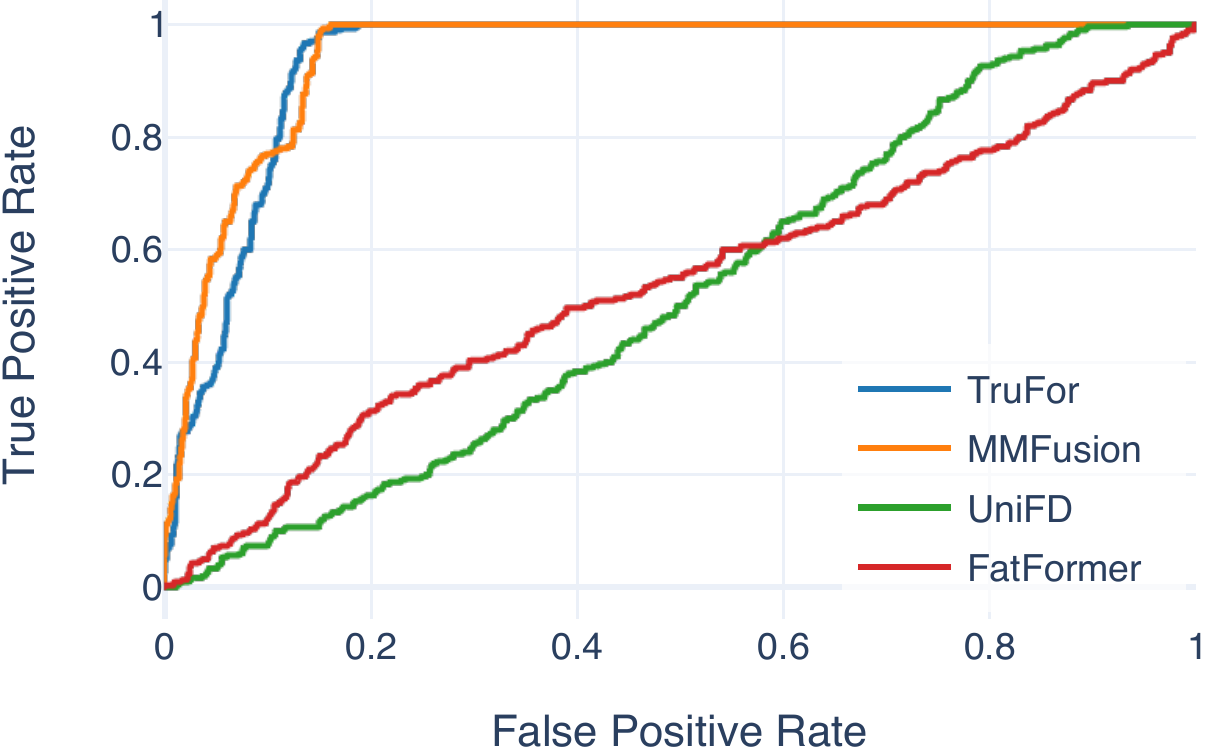}}~
\subfloat[\textit{Attack-1} protocol]{\includegraphics[width=0.45\textwidth]{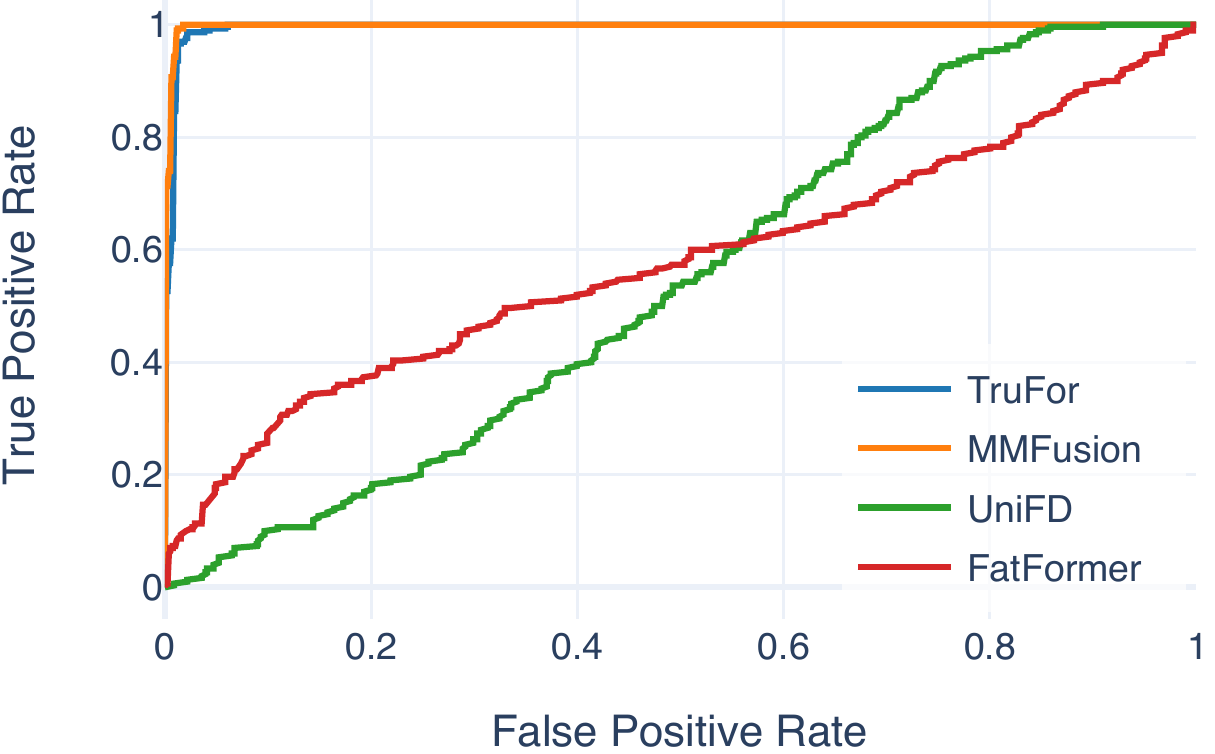}}\\
\subfloat[\textit{Attack-2} protocol]{\includegraphics[width=0.45\textwidth]{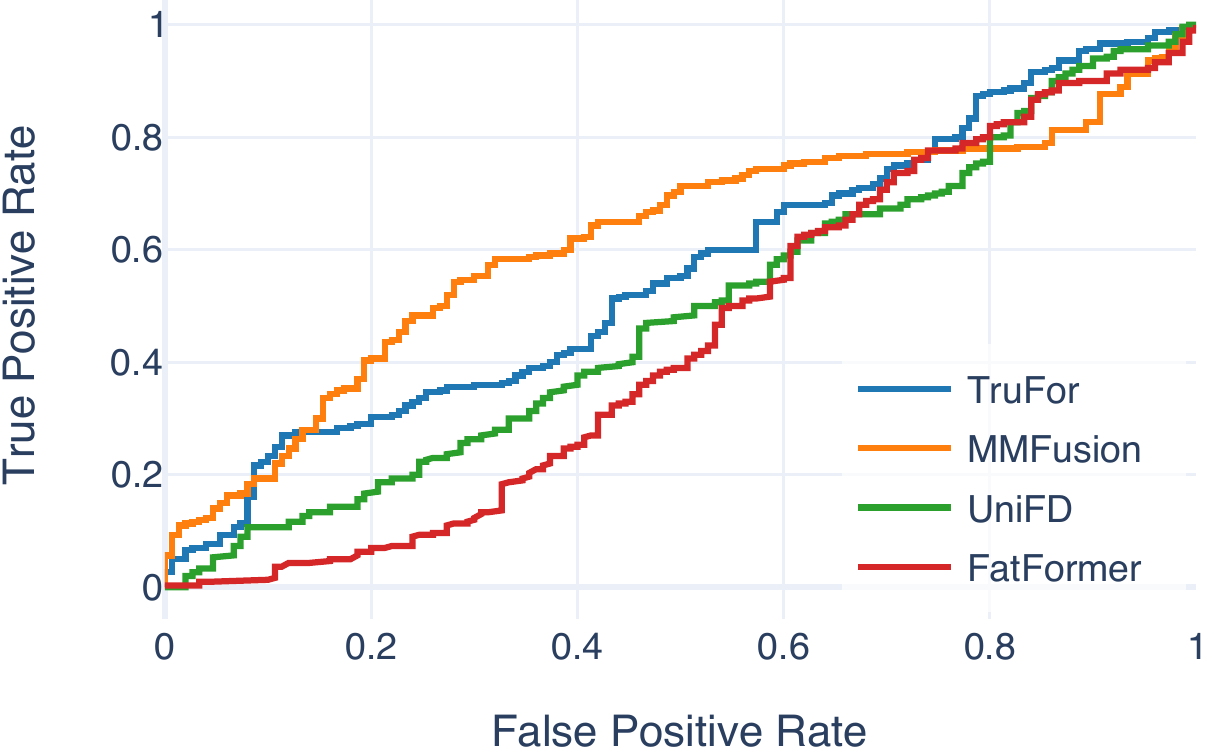}}~
\subfloat[\textit{Attack-3} protocol]{\includegraphics[width=0.45\textwidth]{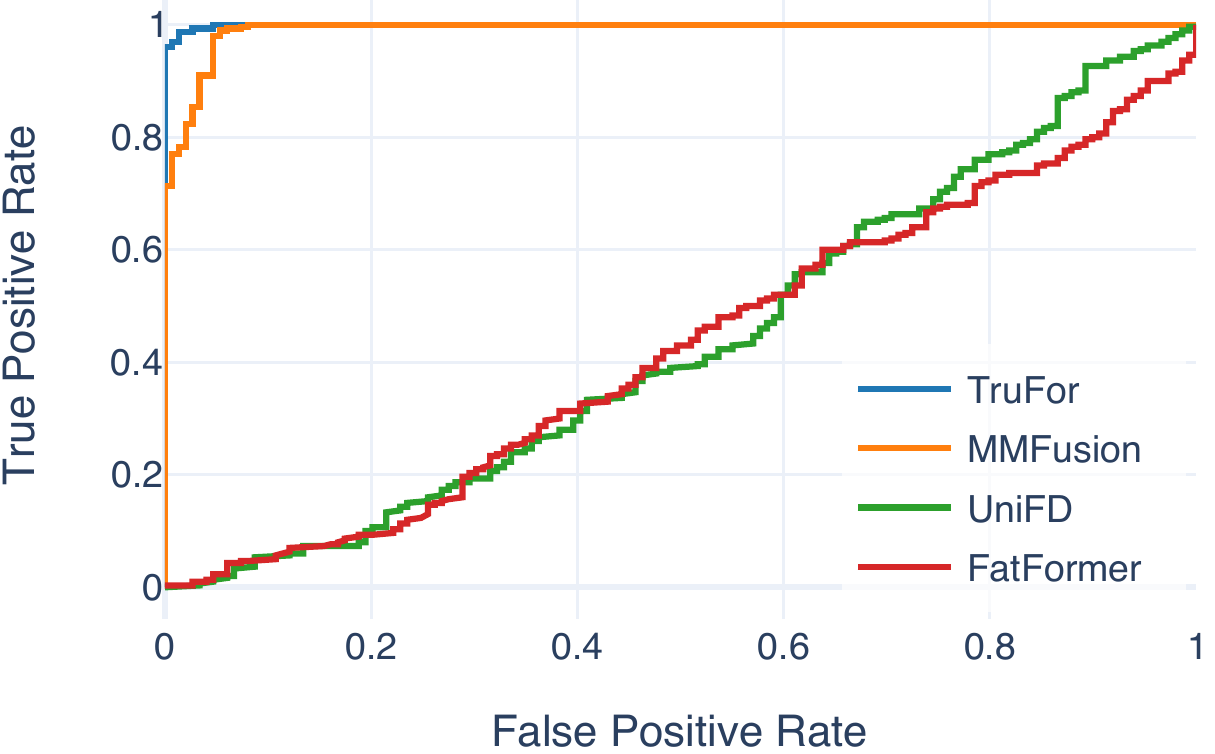}}
\caption{ROC plots of the baseline detection methods on all samples and on each of the attacks from the test set of FantasyID.}
\label{fig:roc-digital}
\end{figure*}


\begin{figure*}[htb]
\centering
\subfloat[\textit{all} protocol]
{\includegraphics[width=0.39\textwidth]{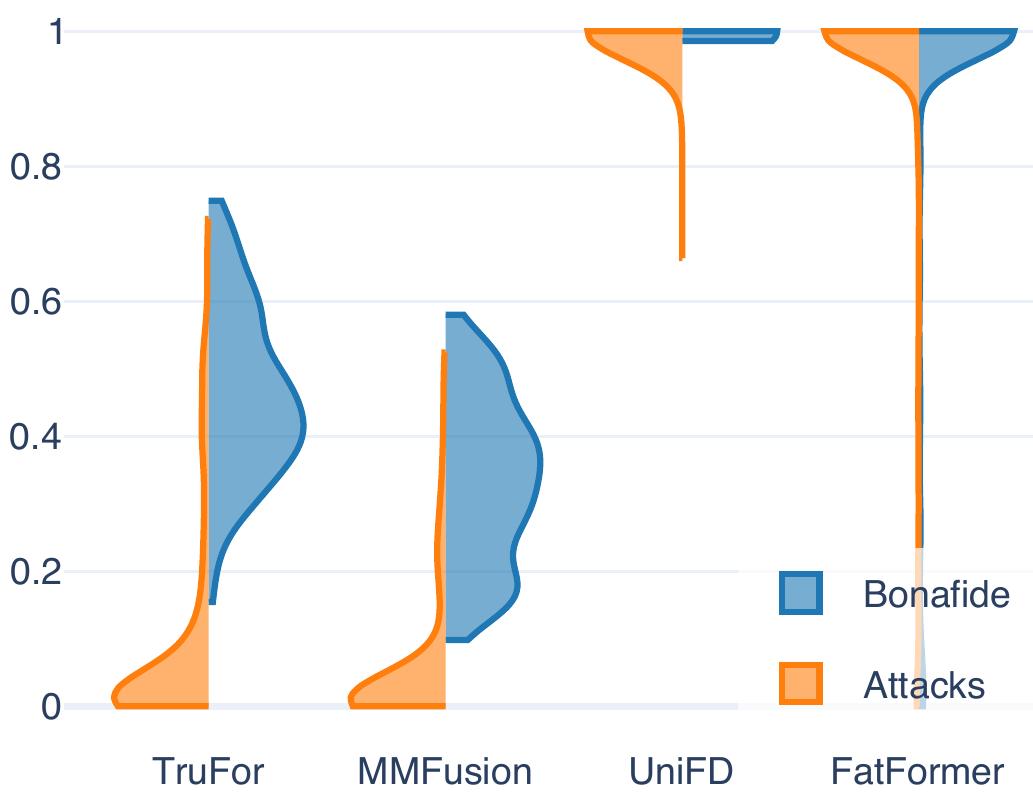}}
~
\subfloat[\textit{Attack-1} protocol]{\includegraphics[width=0.39\textwidth]{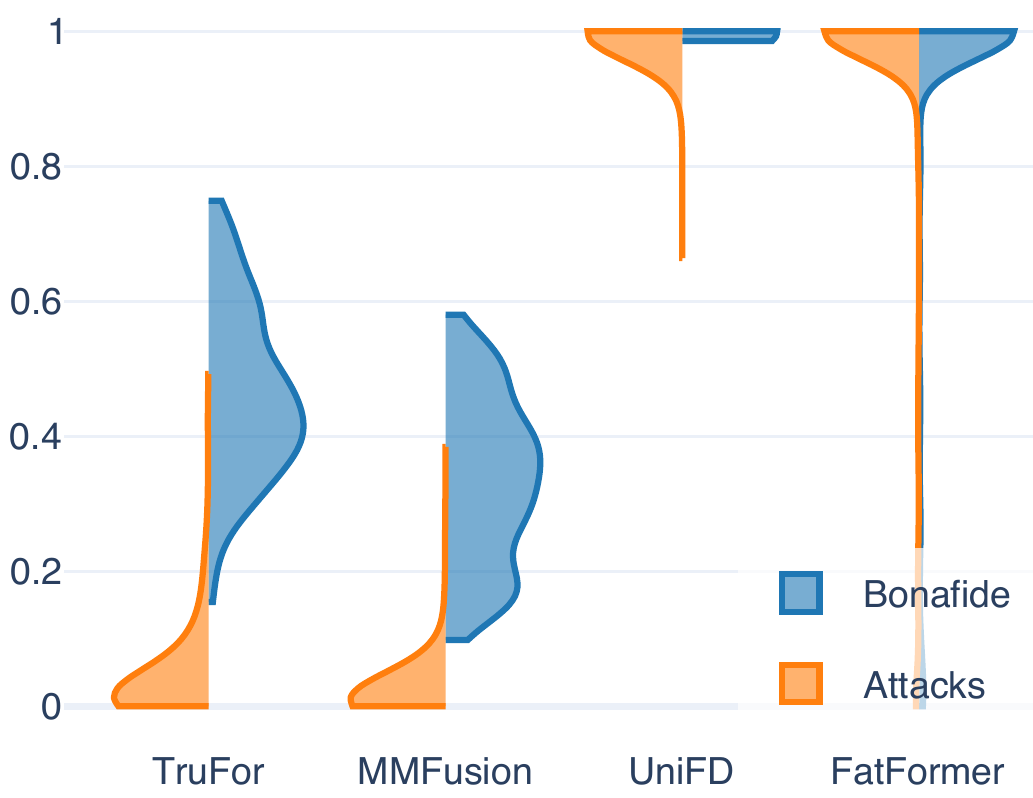}}\\
\subfloat[\textit{Attack-2} protocol]{\includegraphics[width=0.39\textwidth]{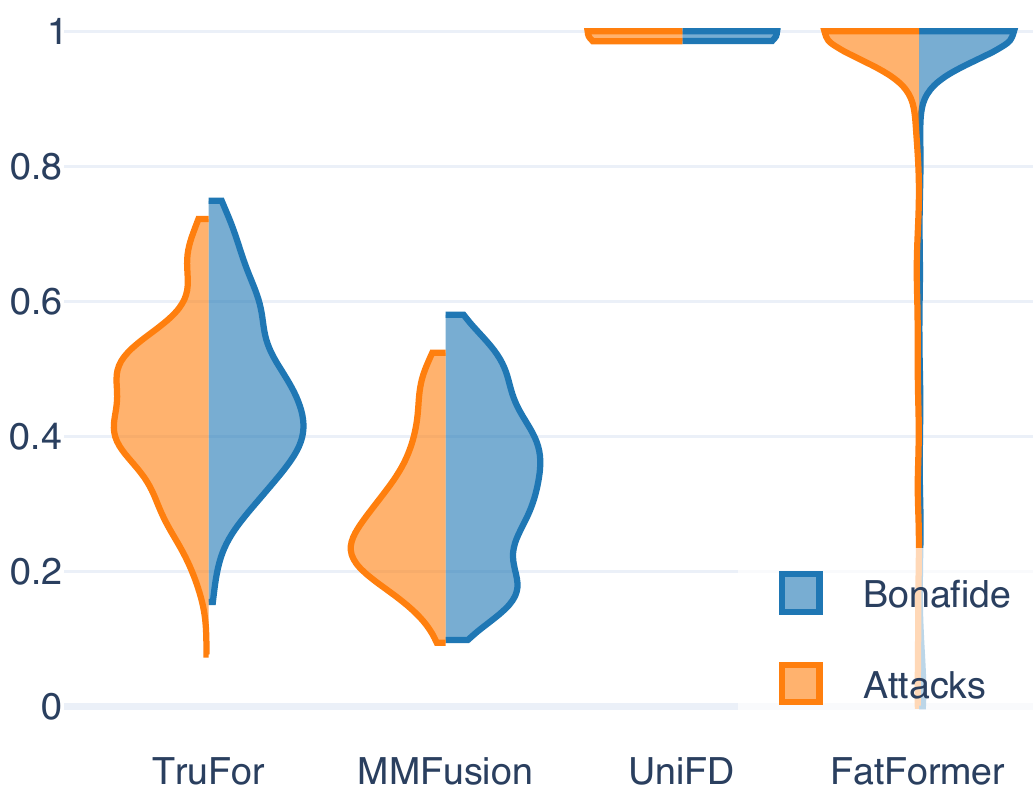}}
~
\subfloat[\textit{Attack-3} protocol]{\includegraphics[width=0.39\textwidth]{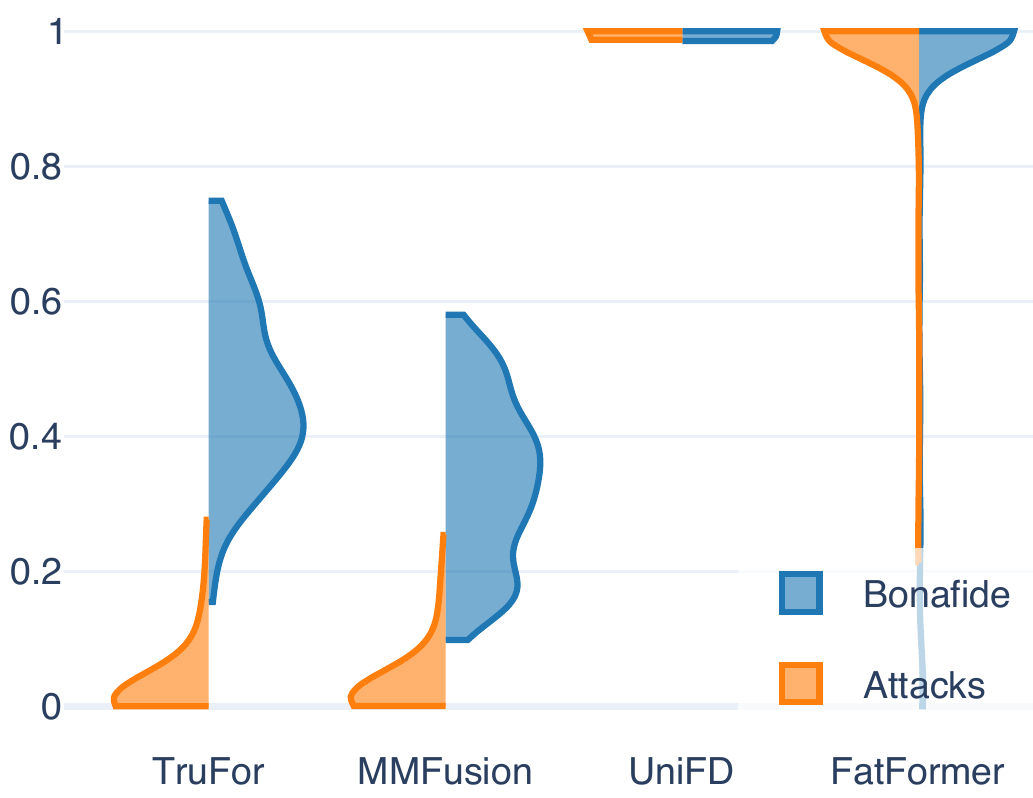}}
\caption{Kernel density estimation (KDE) plots of scores from the baseline detection methods for bonafide and different attacks, including all attacks, Attack-1,3 (text manipulations) and Attack-2 (face manipulation). }
\label{fig:volin_plot}
\end{figure*}

We evaluated manipulation detection algorithms in terms of their binary detection performance. The evaluation is done separately for different digital manipulations, i.e, Attack-1, Attack-2, Attack-3, and \emph{all} representing aggregated result of the three attacks. 

\cref{tab:results-digital} shows the results of digital manipulation detection by our four baseline algorithms using metrics defined in Section~\ref{sec:protocol}. The results clearly demonstrate the advantage of TruFor~\cite{guillaro_trufor_2023} and its extension MMFusion~\cite{rudinac_exploring_2024}, compared to CLIP-based FatFormer and UniFD methods. HTER and AUC metrics show that MMFusion is better at detecting several small manipulated text regions (Attack-1,3), with $HTER=23.9\%$ and $AUC$ above $99\%$. TruFor falls behind with $HTER=31.1\%$ in the detection of manipulated text regions. Both FatFormer and UniFD show near random performance with average $HTER=50\%$ on Attack=1,3.  Their poorer performance can be attributed to the resizing operation (input image is $224\times 224$) which attenuates the manipulation artifacts around small text regions.

While the performance is impressive for text manipulation detection, all the baselines fail to detect manipulation when only faces are edited, i.e, Attack-2. MMFusion performs better than random with $HTER=37.8\%$, followed by TruFor, UniFD and Fatformer with $HTER=48.3\%,50.0\%,58.2\%$, respectively. Attack-2, is created using Facedancer~\cite{rosberg2023facedancer} which blends the facial regions with its background with a gaussian blur. Whereas, all the text manipulations are with simple alpha blending, hence creating a cut-paste attack. TruFor~\cite{guillaro_trufor_2023} is known to work well for these cut-paste manipulations, hence we observe higher performance on text manipulations and not on faces.

TruFor and MMFusion models treat forgery detection as a binary localization problem and have been trained to detect fake images where only some parts have been tampered with, a scenario similar to our digital manipulation protocol. Whereas, UniFD and FatFormer were trained to detect fully generated images. This is reflected in both ~\cref{tab:results-digital} and~\cref{fig:roc-digital}. ROC plots~\cref{fig:roc-digital} show clear dominance of TruFor and its extension MMFusion.

~\cref{fig:volin_plot} shows score distributions of different attacks and 
bonafide for all the baselines. These plots support~\cref{fig:roc-digital} and clearly show Attack-2 (face manipulations only) as the most challenging to detect, since all the baselines have highly overlapping scores for bonafide and manipulated images. The scores for Attack-1,3 show clearer separation with MMFusion and TruFor whereas UniFD and FatFormer fail to distinguish between the bonafide and attacks. The plots illustrate the challenging nature of FantasyID with its diverse attacks.

Overall results demonstrate that even though such state-of-the-art algorithms as MMFusion and TruFor are able to detect the presence of local digital forgeries (mostly text) with a reasonable accuracy, their performance is far from practical applications standards. Since these attacks are easy to perform for a malicious actor, they continue to pose a serious threat to the detection systems.

%% file: sections/conclusion.tex
\section{Conclusion}

In this paper, we presented the first publicly available with a permissive usage license dataset of fantasy ID documents that contain truly bonafide versions of the ID cards (in digital and printed/captured forms) and the fake cards with face and text manipulated. We tested the state-of-the-art forgery detection algorithms on both types of manipulations, demonstrating that binary detection of the forgeries are challenging for the current algorithms. An important future direction is the evaluation of detection algorithms  localization performance, since often a forgery of an ID documents pertains only to small text regions such as a digit of an expiry date, making detection of such manipulations even more challenging.

We believe that this dataset, especially since it is publicly available also for commercial use, will help advance the research in forgery detection and localization in ID documents with the aim to protect users when they are onboarding in many popular online KYC services.